\documentclass[conference]{IEEEtran}
\usepackage{blindtext, graphicx}
\usepackage{comment}
\usepackage{array}
\usepackage{url}
\usepackage{listings}
\usepackage{multirow}
\usepackage{subfig}
\usepackage{color}
\usepackage{booktabs}

\usepackage{amsfonts}
\usepackage[boxed,ruled,vlined]{algorithm2e}

\usepackage{cite}

\ifCLASSINFOpdf
\else
\fi
\usepackage[cmex10]{amsmath}
\usepackage{algorithmic}
\begin{document}
\def \app {Fuse-NILM}

\def \apps {Fuse-NILM~}
%
\title{How good is good enough? Re-evaluating the bar for energy disaggregation}

\author{\IEEEauthorblockN{Nipun Batra}
\IEEEauthorblockA{\small IIIIT-Delhi\\
\small Email: nipunb@iiitd.ac.in}
\and
\IEEEauthorblockN{Rishi Baijal}
\IEEEauthorblockA{\small IIIT-Delhi\\
\small Email: rishi12084@iiitd.ac.in}
\and
\IEEEauthorblockN{Amarjeet Singh}
\IEEEauthorblockA{\small IIIIT-Delhi\\
\small Email: amarjeet@iiitd.ac.in}

\and
\IEEEauthorblockN{Kamin Whitehouse}
\IEEEauthorblockA{\small Univeristy of Virginia\\
	\small Email: whitehouse@virginia.edu}

}


%



\maketitle

\begin{abstract}
Since the early 1980s, the research community has developed ever more sophisticated algorithms for the problem of energy disaggregation, but despite decades of research, there is still a dearth of applications with demonstrated value. In this work, we explore a question that is highly pertinent to this research community: how good does energy disaggregation need to be in order to infer characteristics of a household? We present novel techniques that use unsupervised energy disaggregation to predict both household occupancy and static properties of the household such as size of the home and number of occupants. Results show that basic disaggregation approaches performs up to 30\% better at occupancy estimation than using aggregate power data alone, and are up to 10\% better at estimating static household characteristics. These results show that even rudimentary energy disaggregation techniques are sufficient for improved inference of household characteristics. To conclude, we re-evaluate the bar set by the community for energy disaggregation accuracy and try to answer the question {\em ``how good is good enough?"}.
\end{abstract}

\IEEEpeerreviewmaketitle

\section{Introduction}
Since the early 1980s, the research community has developed ever more sophisticated algorithms for the problem of {\em  energy disaggregation}: the process of breaking down the overall household energy consumption into constituent appliances~\cite{hart_1992}. Many of the basic problems in the field have long been solved, and remaining challenges today include appliances with complex state (such as washing machines), appliances with low power consumption (such as light bulbs), and eliminating the need for extensive training data. These problems are technically challenging, but despite decades of research, there is still a dearth of applications for energy disaggregation that have demonstrated real value or energy savings. As such, many in the community are wondering \emph{``how good is good enough?"} At what point should energy disaggregation be considered a solved problem?

In this work, we explore a question that is highly pertinent to this research community: how good does energy disaggregation need to be in order to infer characteristics of a household? Previous work has shown that aggregate household power readings can be used to estimate dynamic values such as real-time occupancy~\cite{kleiminger2013occupancy, chen2013non} as well as static values such as the size of the home and the number of occupants~\cite{beckel2012towards}. These characteristics can be used for targeted energy feedback~\cite{kleiminger2013occupancy,beckel2012towards}, control of the home's thermostat for improved energy efficiency and user comfort~\cite{pisharoty05thermocoach,lu2010smart}, or targeted energy consulting~\cite{beckel2013automatic}. In this paper, we explore whether disaggregated energy values can either improve the accuracy of this inference or eliminate the need for training data and, if so, how good energy disaggregation must be to do so.

First, we present novel techniques that use unsupervised energy disaggregation to predict household occupancy. Our key intuition is that occupants interact with appliances causing events in the power stream while background loads such as refrigerator operate independently of occupants. Thus, we disaggregate the periodic background loads and predict occupancy based on the remaining aggregate traces, without any further disaggregation. Results show that our unsupervised approach performs up to 30\% better than previous unsupervised approaches that rely on aggregate power data. Additionally, it even performs competitively with supervised techniques that use aggregate power data. 

Second, we present techniques that use unsupervised energy disaggregation to predict six static household properties: age of the home, size of the home, household income, number of floors, number of rooms, and number of occupants. We show that, by disaggregating the heating and cooling power signal alone, classification accuracy increases by up to 10\% over the state-of-the-art. These results show that even rudimentary energy disaggregation techniques are sufficient for improved inference of  household characteristics.
Our results also show that the unsupervised algorithm we use performs worse than the supervised algorithm on traditional NILM metrics, but, better on classifying household characteristics.

To conclude, we re-evaluate the bar set by the community for energy disaggregation accuracy. The techniques we used rely on solutions to classical NILM problems and did not rely on solutions to the difficult challenges that remain today, such as complex loads, low-power loads, and training data. Indeed, our results indicate that techniques that perform poorly in terms of traditional disaggregation accuracy can still perform well when used to infer characteristics of a home. These results indicate a need to introduce application-oriented metrics instead of relying on the traditional metrics of energy disaggregation accuracy. In other words, the community must begin to view energy disaggregation through the lens of applications in order to answer the question {\em ``how good is good enough?"}.

\section{Background and Related Work}
In the early 1980s, George Hart presented the seminal work on energy disaggregation, or non-intrusive load monitoring (NILM)~\cite{hart_1992}. The work was motivated towards providing utilities a platform for residential load research at a large scale. For the end users, energy disaggregation would: 1)  allow demand response via deferring loads, 2) provide detailed energy breakdown to end users, and 3) identify faulty appliances and other power related anomalies. There are three broad directions of work in the community- energy disaggregation methods and traditional metrics, data fusion for better NILM accuracy, and applications of energy disaggregation, all of which we discuss now.

\subsection{Energy disaggregation techniques and traditional metrics}
Smart meter roll outs and the availability of public data sets such as REDD~\cite{redd} have led to a renewed interest in the field in the recent years. A number of disaggregation algorithms have been proposed in the literature in the last three decades, especially in the last few years. These approaches can broadly be categorised into supervised and unsupervised\footnote{\url{http://blog.oliverparson.co.uk/2015/05/what-even-is-supervisedunsupervised.html}}. Parson et al.~\cite{parson_2012} propose unsupervised techniques to iteratively disaggregate loads based on prior models of general appliance types. Kolter et al.~\cite{kolter_2012} propose methods for approximate inference in additive  factorial hidden Markov (AFHMM) based energy disaggregation methods. Beyond the low frequency approaches such as the ones mentioned above, research~\cite{electrisense} has looked into high frequency features (several kHz or more) to detect appliances consuming low power such as electronic devices. 

Traditionally, NILM metrics are chosen based on the \emph{type}\footnote{\url{http://blog.oliverparson.co.uk/2013/12/accuracy-metrics-for-nialm.html}} of problem NILM is being viewed as. Event based metrics assess the accuracy of appliance state changes and are measured using well known machine learning metrics such as F-score, precision and recall. Non-event based metrics assess how well algorithms predict the power demand of individual appliances over time. Examples of such metrics include normalised error in power consumption and percentage of energy correctly allocated, which measures how accurately the disaggregated appliance pie-chart matches with the ground truth pie-chart. 
\subsection{Data fusion to improve NILM accuracy}
Disaggregation in real homes lead to several challenges such as the presence of similar appliances, multiple instances of same appliances and presence of low power electronics. Recent work has looked into fusing data from several pervasive sensors to improve NILM accuracy for such cases. Pathak et al.~\cite{pathak2015acoustic} and Saha et al.~\cite{saha2014energylens} discuss the potential of using acoustic and user WiFi data to localise appliance usages. On similar lines, Akshay et al.~\cite{sn2015loced} improve the accuracy of existing NILM approaches by leveraging the relationship between a user's location as obtained by WiFi localisation and appliance usage. The key idea in such approaches is to constrain the search space of the disaggregation problem by adding contextual information, such as the physical location of loads in the home.

\subsection{Applications of energy disaggregation}
The focus of energy disaggregation by and large has been to provide a detailed energy breakdown to the end users. Prior literature argues that providing appliance level information to end users can help them reduce their energy consumption by as much as 15\%~\cite{darby_2006}. However, recently the applicability and utility of energy disaggregation has been questioned. The fundamental assumption that energy breakdown allows sustained energy savings remains to be validated. Recent work has made an attempt towards answering this question. Parson et al.~\cite{parson_2014} use unsupervised energy disaggregation algorithms to infer fridge usage for 117 homes in the UK. Their goal is to give feedback on the energy-money trade-offs of shifting to new energy-efficient fridges. Barker et al.~\cite{barker2014nilm} also argue that applications such as device scheduling need to be emphasised above plain energy disaggregation accuracy. They find that applications often involve real time constraints which a subset of NILM algorithms fail to adhere to. Recently, Alcala et al.~\cite{alcala2015detecting} discuss the potential of discovering activities of daily living targeted towards healthcare applications using energy disaggregation. Batra et al.~\cite{batra_buildsys_2015} show that while energy disaggregation has the potential to provide targeted feedback to end users for their fridge and HVAC energy consumption, current approaches perform poorly in enabling such feedback. While our work falls in the same category, we question the very nature of the problems that the community needs to focus on.

\section{Predicting  household occupancy}

\begin{figure}[!htb]
	\centering
	\includegraphics[scale=1]{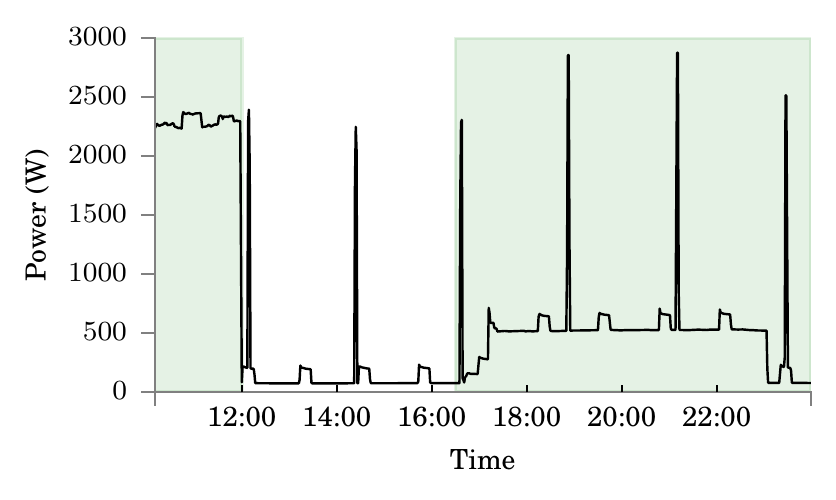} 
	
	\caption{Electricity consumption is often correlated to occupancy and can thus be used to detect occupancy (shown in green shade) in a home.}
	\label{fig:occupancy} 
\end{figure}

Various smart home applications leverage household occupancy for optimised HVAC and light controls~\cite{balaji2013sentinel, agarwal2010occupancy, erickson2011observe}. Occupancy information has also been used for occupant tracking and inferring other activities of daily living towards healthcare applications. Occupancy in homes is traditionally monitored using sensors such as passive infrared (PIR), CO$_2$, etc. However, these sensors are intrusive in nature. Further, previous work highlights that residential deployments scale poorly in terms of cost and effort~\cite{iawe, hnat2011hitchhiker}. 
As a viable alternative, previous research has proposed inferring occupancy from household electricity meter. Since smart meters are rolled out by the utilities and installed outside the home, they scale both in cost and deployment effort and are non-intrusive in nature. Figure \ref{fig:occupancy} shows the relationship between electricity and occupancy of a home, which consumes more power when occupied. We now briefly discuss two existing approaches which extract features from the household aggregate power trace to predict occupancy.

\subsection{Existing approaches}
The basic intuition behind using electricity data for predicting occupancy is that when occupants are present in the home they often interact with appliances, impacting the electricity consumption. Previous approaches by Kleiminger et al.~\cite{kleiminger2013occupancy} and Chen et al.~\cite{chen2013non} use similar notion of finding features in the household aggregate power trace that are indicative of occupancy. Both these algorithms discuss occupancy prediction during non-night hours. We now discuss these two algorithms in more detail.
\subsubsection{Kleiminger et al.~\cite{kleiminger2013occupancy}} compute features such as mean and standard deviation of power over a time window to predict occupancy. High mean usually indicates appliance usage by occupants. High standard deviation indicates usage of appliance with varying power (such as television or laptop) or frequent appliance switch events, both of which indicate occupancy. Kleiminger et al. use standard machine learning classifiers in the supervised settings for predicting occupancy.
\subsubsection{Chen et al.~\cite{chen2013non}} use power range (difference of maximum and minimum power observed in a time window), in addition to the features used by Kleiminger et al. However, unlike the work by Kleiminger et al., their work is unsupervised in nature. Their algorithm consists of two steps. In the first step they compute power range, standard deviation and mean power in the night hours. For the next day, they signal occupancy if any of these three features is above the maximum of the night time value.

\begin{figure*}[!htb]
	\centering
	\includegraphics[scale=1]{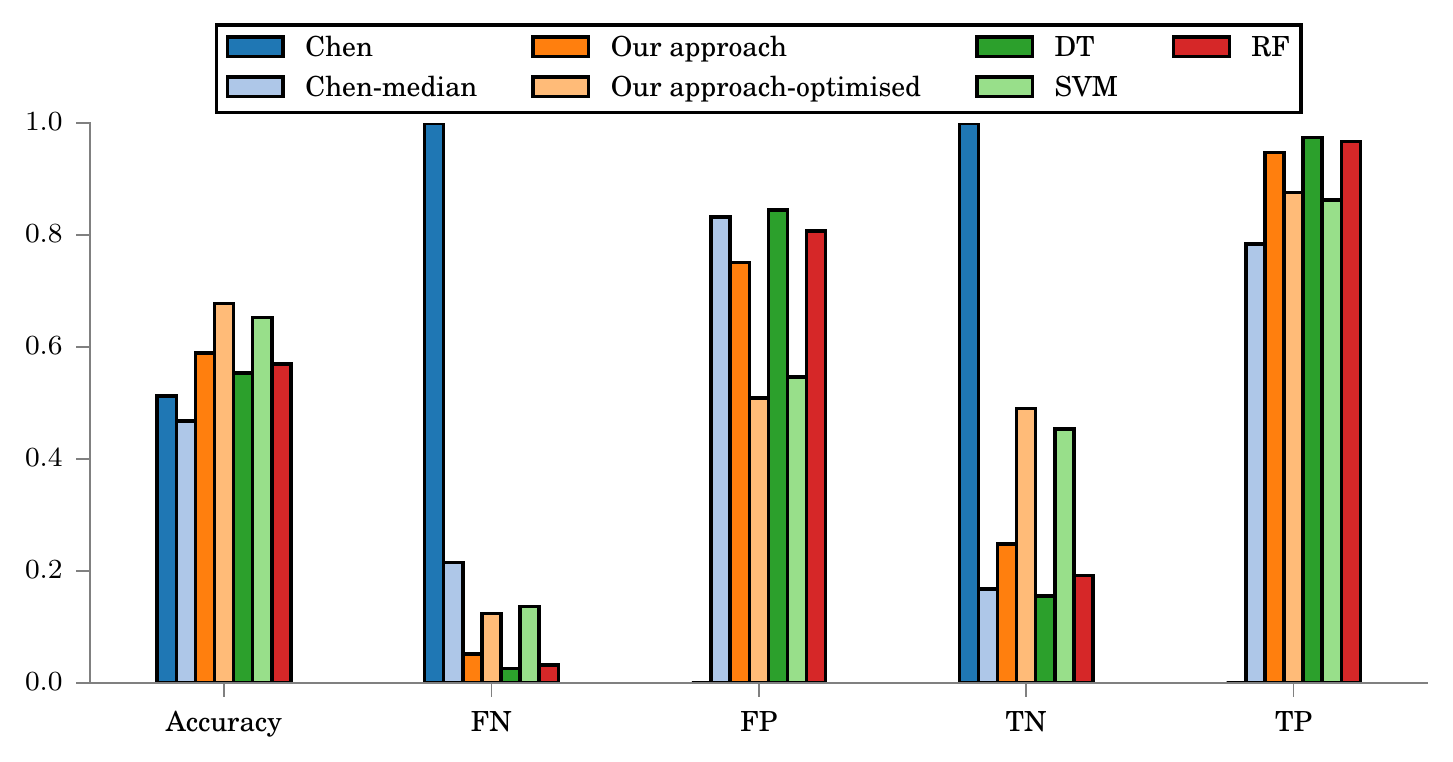} 
	\vspace{-10pt}
	\caption{On our dataset, our unsupervised energy disaggregation based approach of identifying occupancy from electricity gives better performance than both existing supervised and unsupervised approaches relying on aggregate data ($\approx$20\% higher accuracy)}
	
	\label{fig:iawe} 
\end{figure*}

\subsection{Our approach}
Our key intuition is that occupants interact with appliances causing events in the power stream while background loads such as refrigerator operate independently of occupants. Thus, we disaggregate the periodic background loads and predict occupancy based on the remaining aggregate traces, without any further disaggregation. It must be noted that we only need to find events corresponding to the background load turning ON and OFF, and we don't really need to find the energy consumption of these loads, as is usually the practice in energy disaggregation research. Our approach is as follows: 
\begin{enumerate}
	\item \textbf{Event detection}: We first find events in the power stream using the unsupervised event detection method in Hart's seminal NILM algorithm~\cite{hart_1992} as implemented in NILMTK~\cite{nilmtk, kelly2014nilmtk}. An event is said to occur when the aggregate power changes beyond a threshold.
	\item \textbf{Background load removal}: Some of these events are from background loads such as fridge and HVAC and are thus not indicative of occupancy. The events corresponding to background loads are high in number and regular in time due to periodic nature of these loads. Furthermore, following prior work~\cite{parson_2012}, these events can be trivially removed by looking at the night time power trace when typically only background loads are used. 
	\item \textbf{Event pairing}: After removing the events due to background appliances, we are left with events caused by occupants. We pair the rising and the falling edges of similar magnitude and close in time to get the ON and OFF times of different appliances.
	\item \textbf{Filtering event pairs}: Due to inherent limitations in NILM event detection algorithms, some of the events may get missed (true negative), or, additional events could be detected (false positives). We can eliminate these false positives and true negatives by filtering out event pairs which are separated by more than a threshold time period.
	\item \textbf{Predicting occupancy}: We signal occupancy between all event pairs which were within a certain time threshold. Further, since homes in the night hours are typically occupied, yet consume less energy due to people sleeping, we also signal occupancy after the last event till the end of the day; and from the start of the day till the first event. 
	
\end{enumerate}

\subsection{Datasets}
We now describe the two data set we use to evaluate the performance of our approach. 

\subsubsection{Our dataset}: We collected aggregate power data at 1 Hz from a single occupant apartment in Delhi for 12 days using EM6400 smart meter. The occupant marked each time she left or entered the apartment to create ground truth occupancy. The occupancy logs were further validated by two researchers independently and were found to be precise.  

\subsubsection{ECO data set}: The ECO data set was collected from 5 households in Switzerland over a period of 8 months across the summer and the winter season. The data set collected appliance level power data at 1 Hz, household aggregate power data at 1 Hz, and occupancy data entered manually by occupants on tablets. Passive infrared (PIR) sensors were also installed on the entrance doors to ensure that the manual annotation is verified. Further, the data set authors ran sanity scripts to clean days where occupancy wasn't entered correctly.

\subsection{Evaluation}

\subsubsection{Experimental setup}
We now evaluate our approach against the two existing approaches. 
For our dataset which only had a single home, we trained using Kleimenger's approach on the first half of the data set and predicted on the second half. Both Chen's and our unsupervised approaches were also evaluated on the second half of the data set. 
For the supervised approaches used by Kleimenger et al. on the ECO data set, we use 5-fold cross validation, where we train on four homes and test on the remaining home. We use K-Nearest neighbours (with k=5), SVM (RBF kernel) and Random Forest (RF) as the three supervised algorithms. Since Chen's and our algorithm are unsupervised in nature, we make predictions without training on any home. The evaluation was done independently for both summers and winters on the ECO dataset as done in previous work. Since both the existing approaches discussed accuracy of occupancy prediction only over non-night hours, we evaluate occupancy between 6 AM and 10 PM. Further, previous work downsampled the data to 15 minute windows for evaluation. A 15 minute window is labelled to be occupied, if there was any occupancy in this 15 minutes. 

\subsubsection{Evaluation metrics}
Our occupancy prediction problem can be considered as a binary classification task where we take occupied as the positive class and unoccupied as the negative class. \textbf{True positives (TP)} indicate instances where classifier correctly identifies when home is occupied. \textbf{True negatives (TN)} indicate instances where classifier correctly classifies when home is unoccupied. \textbf{False positives (FP)} and \textbf{False negatives (FN)} indicate instances where classifier wrongly signals occupancy and unoccupancy respectively. \textbf{Accuracy} is the percentage of instances whose occupancy status is identified correctly. The scenario in which such a system for occupancy prediction is likely to be used is HVAC control~\cite{scott2011preheat}. Thus, in addition to classification accuracy, we discuss metrics pertinent to HVAC control. The \textbf{energy consumption} of the HVAC is proportional to the time the HVAC is run. We take the simplifying assumption that the HVAC is run for the time when the home is predicted to be occupied (TP+FP). Also, the occupant discomfort which is caused when the home is predicted to be unoccupied when it is occupied, can be taken to be proportional to FN. This metric is often called \textbf{miss time} in the literature. An ideal occupancy prediction system should have low energy consumption and low miss time.

\begin{figure}[!htb]
	\centering
	\includegraphics[scale=1]{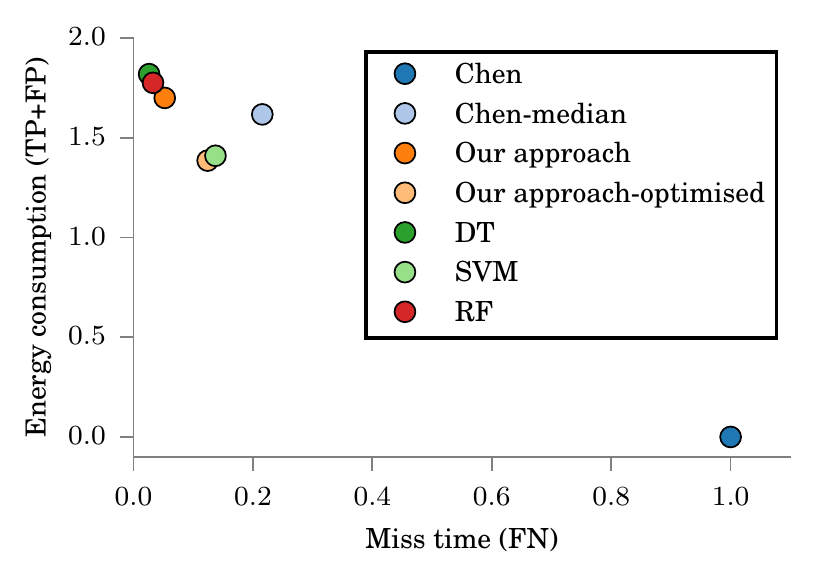} 
	
	\caption{Our unsupervised approach of identifying occupancy from electricity leads to lesser energy consumption with respect to supervised approaches at the cost of slight occupant comfort in comparison to supervised approaches. Chen's approach on the other hand mostly predicts the home to be unoccupied and thus has low energy consumption and more occupant discomfort.}
	\label{fig:iawe_scatter} 
\end{figure}

\subsubsection{Results}

On our dataset, our approach performs better than existing supervised and unsupervised work as shown in Figure \ref{fig:iawe}. Chen's algorithm mostly predicts the home to be unoccupied. This is due to the fact that homes in India typically consume maximum amount of energy in the night time when the air conditioners are on. Thus, the thresholds learnt during night time are poor indicators of baseline energy usage. We also evaluate a variation of Chen's algorithm (called Chen-median) where median value of the features during night time is taken instead of the maximum values. Choosing median features improves the TP. We found that the occupant would typically leave the home around midnight and return after 6 AM. Thus, our approach would wrongly predict the home to be occupied till the first event is seen late in the morning. We evaluate a variation of our approach called `Our-approach optimised' which does not mark the home occupied in the early morning hours. 
We investigated the causes of drop in accuracy for our approach and found that the water heater, which was learnt to be a background load was sometimes used as a foreground load. Our approach currently does not differentiate between usages of such loads which can also be actively used. Previous work has shown that fridges can show an increase in their on durations when actively used. We believe that the cycle duration of background appliances can be used to decide whether they are being actively used or not and leave this to future work. 
While evaluating our approach from the perspective of HVAC control, we observed from Figure \ref{fig:iawe_scatter} that our approach has both less  ``miss time" and less ``energy consumption" in comparison to SVM and Chen-median. Since other supervised approaches mostly predict the home to be occupied, they have low ``miss time", but high ``energy consumption" and are thus not very useful.

\begin{figure*}[!htb]
	\centering
	\includegraphics[scale=1]{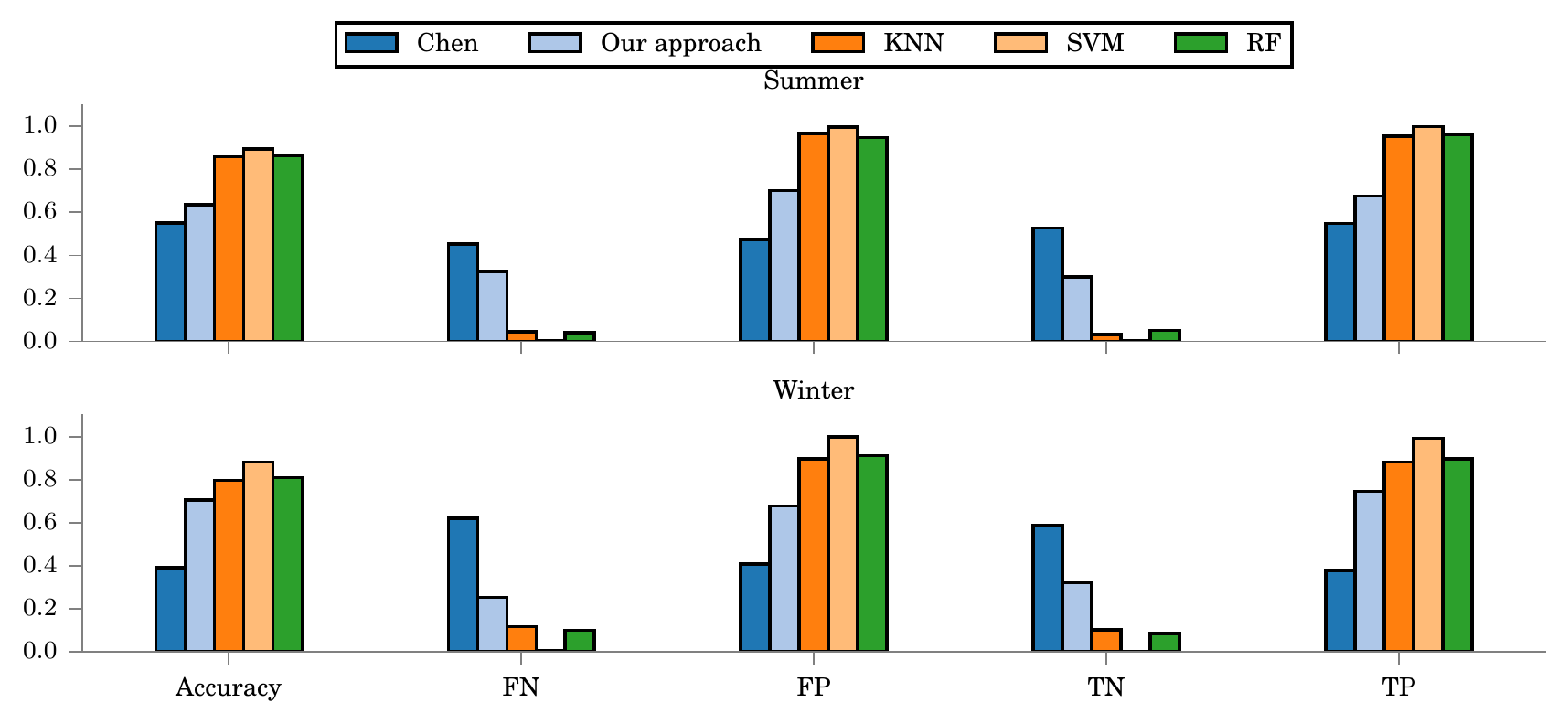} 
	
	\caption{Our unsupervised approach of identifying occupancy from electricity gives better performance on the ECO data set (8\% and 30\% better accuracy in summer and winters respectively) than Chen's unsupervised approach. The supervised approaches tend to always predict the home to be occupied and thus have low FN, but high FP (~30\% higher than ours).}
	
	\label{fig:eco-accuracy} 
\end{figure*}

\begin{figure*}[!htb]
	\centering
	\includegraphics[scale=1]{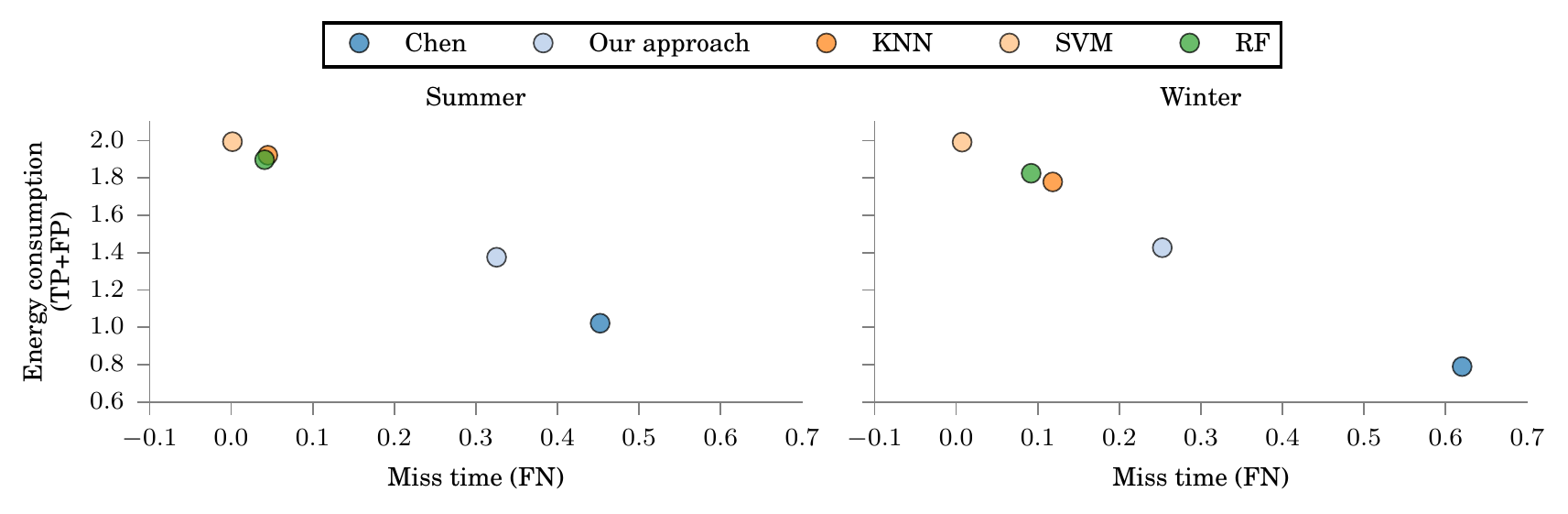} 
	\vspace{-10pt}
	\caption{Our unsupervised approach of identifying occupancy from electricity leads to lesser energy consumption wrt supervised approaches at the cost of occupant comfort. Chen's approach on the other hand mostly predicts the home to be unoccupied and thus has low energy consumption and more occupant discomfort.}
	\label{fig:eco_scatter} 
\end{figure*}

\begin{figure}[!htb]
	\centering
	\includegraphics[scale=1]{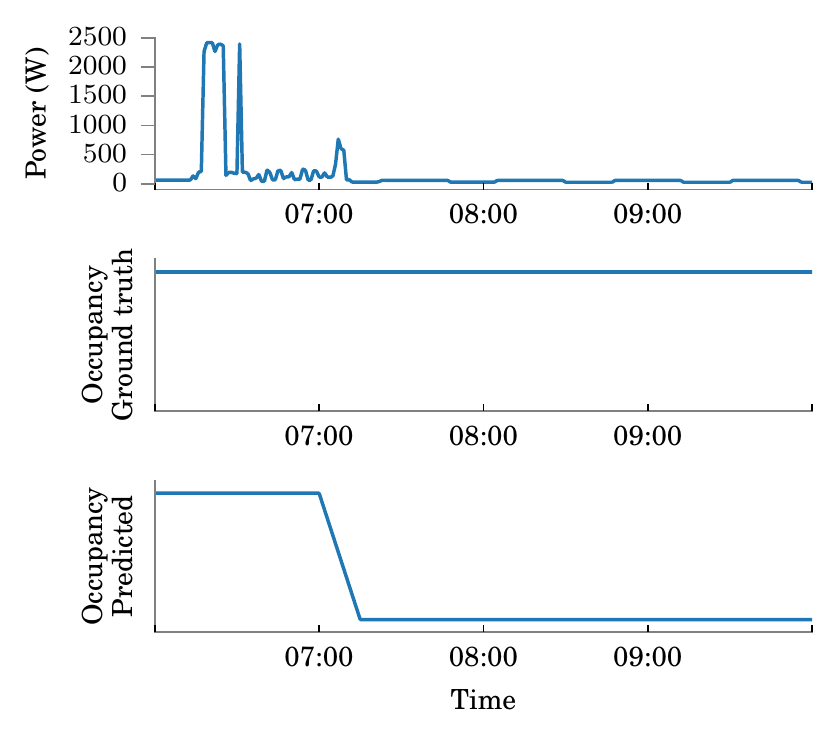} 
	\vspace{-8pt}
	\caption{There are several instances when occupied homes don't use any electricity other than background loads. Such instances cause our occupancy prediction algorithm to wrongly predict home to be unoccupied. }
	\label{fig:eco_low_accuracy} 
\end{figure}

We find from Figure \ref{fig:eco-accuracy}, that our approach gives better accuracy across both summers and winters on the ECO data set in comparison to the unsupervised Chen's approach. We have averaged all the metrics across the five homes in the data set. Chen's approach has a high TN and beats all other approaches on TN. However, it also has a very high FN, signifying that it mostly predicts the home to be unoccupied. On the other hand, supervised algorithms beat our's and Chen's algorithm on TP. Since these homes are mostly occupied, the supervised algorithms have a overall high accuracy as well. It must be taken with a grain of salt that predicting the home to be always occupied will give a high TP and high accuracy given that these homes were occupied most of the times. 
This accuracy obtained by predicting the home to be almost always occupied leads to a high energy consumption as shown in Figure \ref{fig:eco_scatter}. Our approach falls between the supervised approaches and Chen's approach for both energy consumption and miss time. Since Chen's algorithm mostly predicts the home to be unoccupied, it produces a high miss time.
Further, we analysed our results and found that our TP was low due to several instances such as the one shown in Figure \ref{fig:eco_low_accuracy}, where only background loads are used even when the home is occupied. We believe that this is a fundamental limitation of any unsupervised method which predicts occupancy from electricity.

\section{Predicting static household characteristics}
By identifying groups of houses with similar properties, utilities can provide them directed feedback for optimising their energy consumption, or premium energy consulting services~\cite{beckel2013automatic}. Such properties include the area of the home or number of occupants~\cite{beckel2012towards}. In this work, we focus on predicting the following six properties of a household: age, area, income, number of floors, number of rooms, and number of occupants. We now discuss the prior work enabling automated classification of household properties using household energy data.

\subsection{Prior work}
The key intuition behind using energy data for predicting household properties is that households exhibit varied energy consumption patterns as per their inherent property. For instance, a household having five occupants is likely to have higher energy consumption than a single occupant home. In prior work Beckel et al.~\cite{beckel2012towards, beckel2013automatic} extract following types of features from electricity data: 1) consumption-mean power, mean weekday power, mean weekend power, etc.; 2) ratios- ratio of morning to noon power, etc.; 3) temporal- proportion of time power is greater than mean, etc.; and 4) statistical properties- variance, etc. The complete list of features used can be found in Table \ref{tab:prior_features}. They train on these features using standard machine learning classifiers such as k-nearest neighbours to predict the household characteristics using the CER data set, which contains household aggregate power consumption measured at 30 minute interval for more than 4000 households from Ireland. Since this data did not contain appliance level power information, we could not use it for our analysis.

\tabcolsep=0.1cm
\begin{table}
	\begin{center}
		\begin{tabular}{l|p{6cm}}
			\hline
			\textbf{Feature category}&\textbf{List of features}\\
			\hline
			Consumption&Mean total, Mean weekday, Mean weekend, Mean day (6 AM to 10 PM), Mean evening (6 PM to 10 PM), Mean morning (6 AM to 10 AM), Mean night (1 AM to 5 AM), Mean noon (10 AM to 2 PM), Max, Min\\ \hline 
			Ratios&Mean over Maximum, Minimum over Mean, Mean morning over Mean noon, Mean evening over Mean noon, Mean noon over Mean total, Mean night over Mean day, Mean weekday over Mean weekend\\ \hline 
			Temporal&Proportion of time power \textgreater mean, Proportion of time power \textgreater 0.5 kW, Proportion of time power \textgreater 1 kW\\ \hline
			Statistical&Variance, Autocorrelation\\
			\hline
		\end{tabular}
	\end{center}
	\caption{Features used in work by Beckel et al.~\cite{beckel2013automatic} for predicting household characteristics from smart meter data.}
	\label{tab:prior_features}
\end{table}

\begin{figure*}[!htb]
	\centering
	\includegraphics[scale=1]{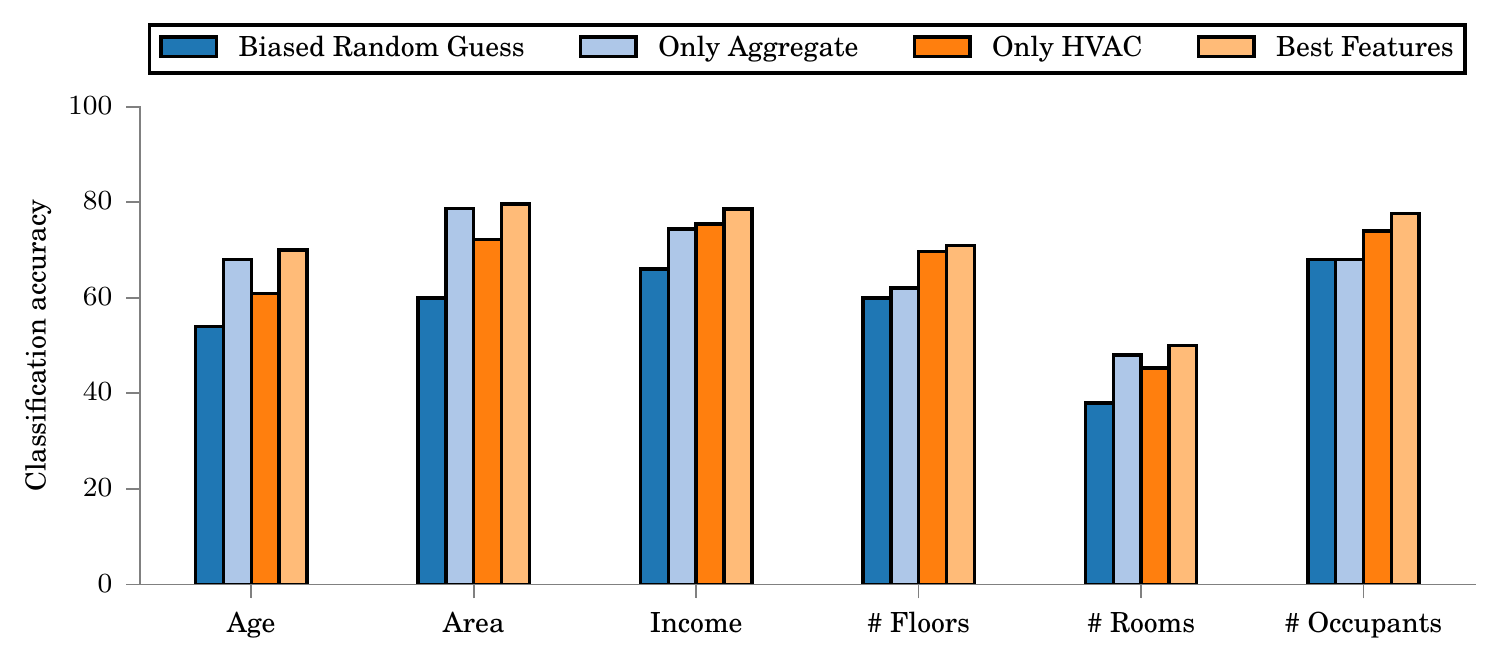} 
	\vspace{-10pt}
	\caption{Our features derived from appliance level power traces can improve the accuracy of classification of income by 4\%,  floors by 9\%, occupants by 9\% and for age, area, rooms by 2\%}
	
	\label{fig:class} 
\end{figure*}

\subsection{Our approach}
Our approach builds upon the work of Beckel et al.~\cite{beckel2013automatic} Our intuition is that there are several additional appliance level features which can help to predict household characteristics more accurately. We believe that HVAC power data is likely to be indicative of household properties since HVAC is typically the largest load in the home and is present across most homes. It must be noted that several other appliances such as refrigerator are also likely to be indicative of household characteristics. However, the data set that we used didn't have refrigerator data for all the homes. All the features in Table \ref{tab:prior_features} were computed on HVAC power stream in addition to aggregate power stream. In addition, we computed the following features:
\begin{enumerate}
	\item \textbf{Maximum HVAC power consumption}: Higher powered HVAC is likely to suggest a higher need for conditioning possibly as a result of high household area or more number of occupants. It could also indicate a household with higher income and high number of rooms in the home.
	\item \textbf{Number of appliance switches}: Higher appliance switches are caused by more interactions which can likely be explained by higher number of occupants.
	\item \textbf{Number of circuits dedicated for HVAC}: It is likely that larger homes (many rooms) or homes spanning multiple floors have multiple circuits dedicated for HVAC.
	\item \textbf{Proportion of time the HVAC is on}: Old homes with insufficient insulation are likely to have high thermal leakages and thus the HVAC may need to remain on for a longer duration.
	\item \textbf{Fraction of energy consumed by HVAC}: may also indicate an old poorly insulated home.
	\item \textbf{Power of highest power consuming appliance (Mean, Max, Median)}: may indicate larger homes, or more occupants. This is found by using Hart's event detection NILM algorithm as discussed earlier for occupancy prediction.
	
\end{enumerate}
We select the $k$ best features for each household characteristic by using univariate feature selection based on chi squared testing~\cite{sklearn-feature}. It must be noted that all features must have non-negative values for this method to hold true. Our aim is to prevent overfitting by reducing the dimensionality of the data set and to understand the relative importance of different features for the classification task.
\subsection{Data set}
We use the publicly available Dataport~\cite{parson2015dataport} data set for our evaluation. Dataport contains appliance and household aggregate power data logged once every minute for more than 700 homes across three years. It also contains household characteristics metadata for a small subset of the homes. In total, there are 50 homes for which household aggregate power data, appliance level power data and household characteristics metadata is available. Table \ref{tab:dataset} shows the distribution of different household characteristics in the Dataport dataset. We have divided the data into different classes keeping in mind: 1) definition of classes used in prior literature~\cite{beckel2012towards}; 2) having roughly equal number of homes across different classes.

\tabcolsep=0.1cm
\begin{table}
	\begin{center}
		\begin{tabular}{ccc}
			\hline
			\textbf{Characteristic}&\textbf{Classes}&\textbf{Number of samples}\\
			\hline
			\multirow{ 2}{*}{Age}& Old ( Age $\ge$ 30 years) & 23\\
			& New (Age $\le$ 30 years ) & 27\\  \hline 
			\multirow{ 2}{*}{Area}& Medium (900 sq. ft $\le$ Area $\le$ 1800 sq. ft) & 20\\
			& High (Area $\ge$ 1800 sq. ft)  & 30\\ \hline
			\multirow{ 2}{*}{Income}&  Below \$150,000/year & 33\\
			& Above \$150,000/year & 17\\ \hline
			\multirow{ 2}{*}{\# Floors}&  One  & 30\\
			& Two or more & 20\\  \hline 
			\multirow{ 3}{*}{\# Rooms}&  $\le$ 6  & 19\\
			& \textgreater 6 and $\le$ 8 & 18\\ 
			&\textgreater 8 & 13\\ \hline 
			\multirow{ 2}{*}{\# Occupants}&  $\le$ 2& 34\\
			& \textgreater 2 & 16\\ \hline

			\hline
		\end{tabular}
	\end{center}
	\caption{Household characteristics from Dataport data set.}
	\label{tab:dataset}
\end{table}

\subsection{Evaluation}
In our evaluation, we first try to answer if the features computed from HVAC  power trace improve the accuracy of predicting household characteristics. Next, we evaluate the accuracy of prediction of HVAC power traces. Finally, we evaluate the accuracy of prediction of household characteristics by using the disaggregated power trace instead of the appliance level power trace.
\subsubsection{Evaluation of household characteristics using appliance level data}
Our experimental setup consisted of doing a two-fold cross validation on the 50 homes from the Dataport dataset. For each household characteristic, we learn the optimum features by using univariate feature selection as described earlier in our approach. To evaluate the value of information added by our HVAC features, we learn the optimal features from: 1) only aggregate power features; 2) only HVAC power features; and 3) from both HVAC and aggregate power features. In addition, we compute the accuracy of biased random guess- i.e. predict the most common class for all the homes. The HVAC features are computed from the appliance level data made available in the data set.

For 4 out of the 6 household properties, we find that the optimal feature set includes the features we proposed (Table \ref{tab:top-features}). Figure \ref{fig:class} compares the accuracy of prediction of household characteristics when different set of features are used. For age and area, the optimal features do not contain HVAC based features and thus the optimal accuracy is obtained from household aggregate power features. However, for income, number of floors, number of rooms and number of occupants, our features improve the accuracy of classification by upto 9\%.

\begin{table}
	\begin{center}
		\begin{tabular}{l|p{6cm}} \hline 
			\textbf{Characteristic}&\textbf{Optimal features}\\
			\hline
			Age&\textbf{Aggregate:} evening mean, max, mean, weekday mean, weekend mean\\ \hline
			Area&\textbf{Aggregate:} max, mean, morning mean, night mean, weekday mean, weekend mean\\ \hline 
			\multirow{2}{*}{Income}&\textbf{HVAC:} max power, night mean \\
			& \textbf{Aggregate:} evening mean, max, mean, night mean, night mean, weekday mean, weekend mean\\ \hline 
			\# floors & \textbf{HVAC:} All HVAC features\\ \hline
			\# rooms& \textbf{HVAC:} max power\\
			 & \textbf{Aggregate:}max, night mean\\ \hline
			 \# occupants& \textbf{Aggregate}:night mean\\
			 & \textbf{Miscellaneous:} mean, max, median of highest power consuming appliance.\\
						
			\hline 
		
		\end{tabular}
	\end{center}
	\caption{For 4 out of 6 household properties, the optimal feature set contains additional energy disaggregation based features we proposed.}
	\label{tab:top-features}
\end{table}

%
%

\subsubsection{Evaluation of energy disaggregation accuracy}
Having found that HVAC data can improve the classification accuracy for household properties, we now evaluate the accuracy of breaking down aggregate data into HVAC data. We compare the accuracy of HVAC disaggregation using unsupervised Hart's algorithm to the well-known FHMM~\cite{kolter_2012} algorithm in the supervised setting. We used the first half for training the FHMM and disaggregate on the same time interval as we do for Hart's algorithm. We use the following standard definition of NILM metrics to evaluate HVAC disaggregation~\cite{nilmtk}:
\begin{enumerate}
	\item \% Error in Energy: $\frac{|\text{Predicted energy - Actual energy}|\times 100\%}{\text{Actual energy}}$
	\vspace{2mm}
	\item Root Mean Squared Error (RMSE) Power:
	
	$ {\sqrt {\frac{1} {N}{\sum\limits_{i = 1}^N {(\text{Predicted power}_i -  \text{Actual power}_i} })^{2} } }$
	
	\vspace{2mm}
	\item F-score: First, disaggregation is converted to a binary classification problem where an appliance is ON if it consumes more than a threshold and OFF otherwise. Next, the standard definition of F-score is used on this binary classification task.
\end{enumerate}

\begin{table}
	\centering

	\begin{tabular}{cccc}
		\hline
		\textbf{Algorithm} & \textbf{Error Energy (\%)} & \textbf{F-score} & \textbf{RMSE (W)} \\ \hline
		Hart      & 55                & 0.66    &      1029    \\
		FHMM      & 47                & 0.8     & 708     \\ \hline
	\end{tabular}
	\caption{Supervised FHMM gives better accuracy than the unsupervised Hart's algorithm for HVAC disaggregation on traditional NILM metrics.}
	\label{nilm-table}
\end{table}

From Table \ref{nilm-table} we can observe that supervised FHMM algorithm gives better disaggregation accuracy across all the traditional NILM metrics. Given these results, we would expect FHMM to give better accuracy than Hart's algorithm in predicting household characteristics.

\begin{figure}[!htb]
	\centering
	\includegraphics[scale=1]{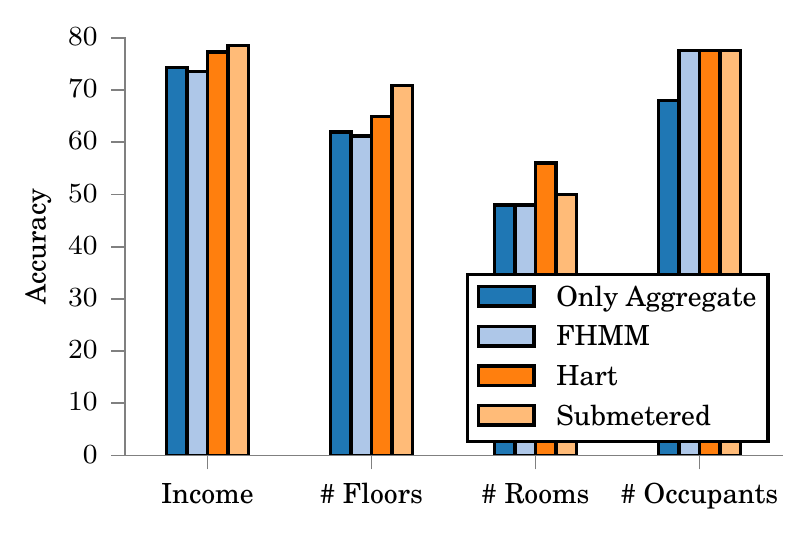} 
	
	\caption{Unsupervised energy disaggregation (Hart's algorithm) produces appliance data with sufficient accuracy to improve classification of household properties over using just aggregate data by 2\%, 3\%, 8\% and 10\% for income, \# floors, \# rooms \# occupants and respectively. It also gives better accuracy than the accuracy obtained from using features based on supervised (FHMM) disaggregated trace.}
	
	\label{fig:nilm_accuracy} 
\end{figure}

\begin{figure}[!htb]
	\centering
	\includegraphics[scale=1]{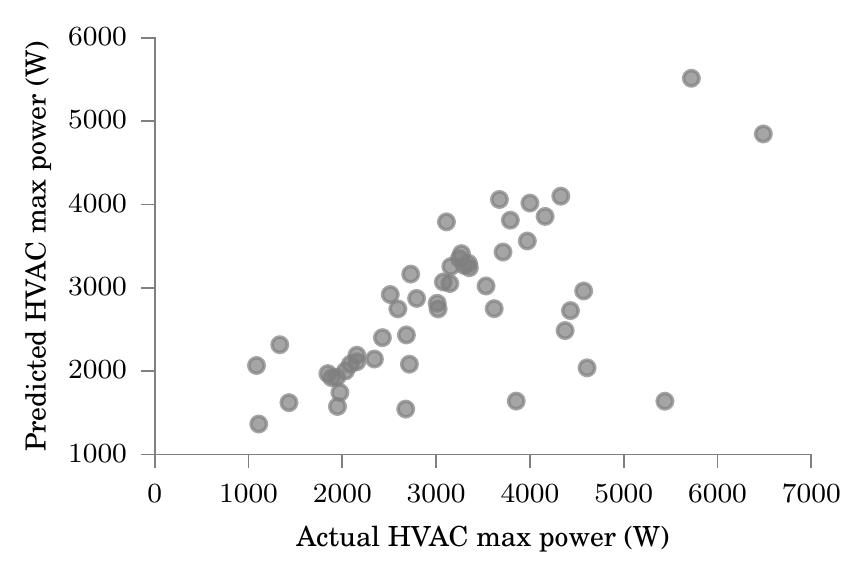} 
	
	\caption{Predicted HVAC max power using Hart's algorithm gives a high correlation of 0.65 with actual HVAC max power usage}
	
	\label{fig:hvac_power_scatter} 
\end{figure}

\begin{figure}[!htb]
	\centering
	\includegraphics[scale=1]{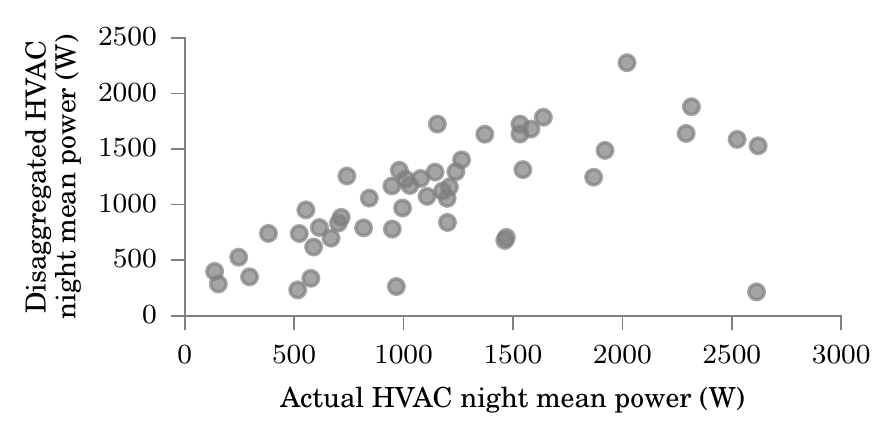} 
	
	\caption{Night time HVAC power is an important feature for predicting household characteristic. It computed on Hart's disaggregated stream correlates highly (R$^2$=0.62) with actual night time HVAC power consumption.}
	
	\label{fig:hvac_night_scatter} 
\end{figure}
\subsubsection{Evaluation of household characteristics using disaggregated data}
Having seen that HVAC disaggregation is far from perfect and the fact that supervised FHMM performs better than unsupervised Hart's algorithm, we evaluate the accuracy of classification of household characteristics on features learnt from disaggregated HVAC data. We limit our evaluation to income, number of floors, number of rooms and number of occupants, as for the other two characteristics, aggregate features give highest accuracy. Surprisingly, we find that for predicting household characteristics, Hart's algorithm gives better performance in comparison to FHMM. For predicting income and number of floors, the accuracy of classification based on the trace predicted by Hart's algorithm, drops in comparison to when appliance level HVAC data was used. However, the accuracy is still 2\% and 3\% better than if only aggregate data was used. For number of rooms and number of occupants, using features learnt from disaggregated traces improves the accuracy by 8 and 10\% respectively. We believe that the high accuracy can be explained by the fact that disaggregation produces accurate power traces corresponding to the features which have a significant impact on classification accuracy. 

Figure \ref{fig:hvac_power_scatter} shows that there is a high correlation between the  `HVAC night mean power' extracted from the appliance level HVAC data and from Hart's disaggregated power trace. Similarly, Figure \ref{fig:hvac_night_scatter} shows that there is a high correlation between `HVAC max power' extracted from appliance level and Hart's disaggregated power trace. Both these features significantly impact classification accuracy and their closeness to actual features learnt from appliance level data explain the high classification accuracy.

\section{Conclusions}
In this work, we explored the question of how accurate must NILM be for it to be applicable in the real world. Our results indicate that disaggregating high energy consuming periodic background loads such as HVAC and fridges in an unsupervised fashion has a significant scope in applications such as prediction of household characteristics. Our results show that basic disaggregation approaches performs up to 30\% better at occupancy estimation than using aggregate power data alone, and are up to 10\% better at estimating static household characteristics. Our results show that even rudimentary energy disaggregation techniques are sufficient for improved inference of household characteristics. We finally conclude that ``how good is good enough" can be answered by looking through the lens of applications.
\section{Limitations and Future work}

Some of the household properties can be considered as continuous variables and instead of treating the problem as a classification problem, we plan to consider the problem as a regression problem. Current work only looked at a month of data for predicting static household characteristics. In the future, we would like to analyse long term trends to 1) improve accuracy be incorporating household's response to weather to improve classification accuracy; 2) indicate variations in these static characteristics and alert homes when they deviate from their baseline.

\bibliographystyle{abbrv}
\bibliography{reference}  

\begin{thebibliography}{10}

\bibitem{sn2015loced}
Loced: Location-aware energy disaggregation framework.

\bibitem{agarwal2010occupancy}
Y.~Agarwal, B.~Balaji, R.~Gupta, J.~Lyles, M.~Wei, and T.~Weng.
\newblock Occupancy-driven energy management for smart building automation.
\newblock In {\em Proceedings of the 2nd ACM Workshop on Embedded Sensing
  Systems for Energy-Efficiency in Building}, pages 1--6. ACM, 2010.

\bibitem{alcala2015detecting}
J.~Alcal{\'a}, O.~Parson, and A.~Rogers.
\newblock Detecting anomalies in activities of daily living of elderly
  residents via energy disaggregation and cox processes.
\newblock 2015.

\bibitem{balaji2013sentinel}
B.~Balaji, J.~Xu, A.~Nwokafor, R.~Gupta, and Y.~Agarwal.
\newblock Sentinel: occupancy based hvac actuation using existing wifi
  infrastructure within commercial buildings.
\newblock In {\em Proceedings of the 11th ACM Conference on Embedded Networked
  Sensor Systems}, page~17. ACM, 2013.

\bibitem{barker2014nilm}
S.~Barker, S.~Kalra, D.~Irwin, and P.~Shenoy.
\newblock Nilm redux: The case for emphasizing applications over accuracy.
\newblock In {\em NILM-2014 Workshop}, 2014.

\bibitem{iawe}
N.~Batra, M.~Gulati, A.~Singh, and M.~B. Srivastava.
\newblock {It's Different: Insights into home energy consumption in India}.
\newblock In {\em Proceedings of the Fifth ACM Workshop on Embedded Sensing
  Systems for Energy-Efficiency in Buildings}, 2013.

\bibitem{nilmtk}
N.~Batra, J.~Kelly, O.~Parson, H.~Dutta, W.~Knottenbelt, A.~Rogers, A.~Singh,
  and M.~Srivastava.
\newblock {NILMTK: An Open Source Toolkit for Non-intrusive Load Monitoring}.
\newblock In {\em Fifth International Conference on Future Energy Systems (ACM
  e-Energy)}, Cambridge, UK, 2014.

\bibitem{batra_buildsys_2015}
N.~Batra, A.~Singh, and K.~Whitehouse.
\newblock If you measure it, can you improve it? exploring the value of energy
  disaggregation.
\newblock In {\em Proceedings of the second ACM International Conference on
  Embedded Systems For Energy-Efficient Built Environments}. ACM, 2015.

\bibitem{beckel2012towards}
C.~Beckel, L.~Sadamori, and S.~Santini.
\newblock Towards automatic classification of private households using
  electricity consumption data.
\newblock In {\em Proceedings of the Fourth ACM Workshop on Embedded Sensing
  Systems for Energy-Efficiency in Buildings}, pages 169--176. ACM, 2012.

\bibitem{beckel2013automatic}
C.~Beckel, L.~Sadamori, and S.~Santini.
\newblock Automatic socio-economic classification of households using
  electricity consumption data.
\newblock In {\em Proceedings of the fourth international conference on Future
  energy systems}, pages 75--86. ACM, 2013.

\bibitem{chen2013non}
D.~Chen, S.~Barker, A.~Subbaswamy, D.~Irwin, and P.~Shenoy.
\newblock Non-intrusive occupancy monitoring using smart meters.
\newblock In {\em Proceedings of the 5th ACM Workshop on Embedded Systems For
  Energy-Efficient Buildings}, pages 1--8. ACM, 2013.

\bibitem{darby_2006}
S.~Darby.
\newblock The effectiveness of feedback on energy consumption.
\newblock {\em A Review for DEFRA of the Literature on Metering, Billing and
  direct Displays}, 2006.

\bibitem{erickson2011observe}
V.~L. Erickson, M.~{\'A}. Carreira-Perpi{\~n}{\'a}n, and A.~E. Cerpa.
\newblock Observe: Occupancy-based system for efficient reduction of hvac
  energy.
\newblock In {\em Information Processing in Sensor Networks (IPSN), 2011 10th
  International Conference on}, pages 258--269. IEEE, 2011.

\bibitem{electrisense}
S.~Gupta, M.~S. Reynolds, and S.~N. Patel.
\newblock Electrisense: single-point sensing using emi for electrical event
  detection and classification in the home.
\newblock In {\em Proceedings of the 12th ACM international conference on
  Ubiquitous computing}, pages 139--148. ACM, 2010.

\bibitem{hart_1992}
G.~W. Hart.
\newblock Nonintrusive appliance load monitoring.
\newblock {\em Proceedings of the IEEE}, 80(12):1870--1891, 1992.

\bibitem{hnat2011hitchhiker}
T.~W. Hnat, V.~Srinivasan, J.~Lu, T.~I. Sookoor, R.~Dawson, J.~Stankovic, and
  K.~Whitehouse.
\newblock The hitchhiker's guide to successful residential sensing deployments.
\newblock In {\em Proceedings of the 9th ACM Conference on Embedded Networked
  Sensor Systems}, pages 232--245. ACM, 2011.

\bibitem{kelly2014nilmtk}
J.~Kelly, N.~Batra, O.~Parson, H.~Dutta, W.~Knottenbelt, A.~Rogers, A.~Singh,
  and M.~Srivastava.
\newblock Nilmtk v0. 2: a non-intrusive load monitoring toolkit for large scale
  data sets: demo abstract.
\newblock In {\em Proceedings of the 1st ACM Conference on Embedded Systems for
  Energy-Efficient Buildings}, pages 182--183. ACM, 2014.

\bibitem{kleiminger2013occupancy}
W.~Kleiminger, C.~Beckel, T.~Staake, and S.~Santini.
\newblock Occupancy detection from electricity consumption data.
\newblock In {\em Proceedings of the 5th ACM Workshop on Embedded Systems For
  Energy-Efficient Buildings}, pages 1--8. ACM, 2013.

\bibitem{kolter_2012}
J.~Z. Kolter and T.~Jaakkola.
\newblock {Approximate Inference in Additive Factorial HMMs with Application to
  Energy Disaggregation}.
\newblock In {\em Proceedings of the International Conference on Artificial
  Intelligence and Statistics}, pages 1472--1482, La Palma, Canary Islands,
  2012.

\bibitem{redd}
J.~Z. Kolter and M.~J. Johnson.
\newblock {REDD: A public data set for energy disaggregation research}.
\newblock In {\em Proceedings of 1st KDD Workshop on Data Mining Applications
  in Sustainability}, San Diego, CA, USA, 2011.

\bibitem{lu2010smart}
J.~Lu, T.~Sookoor, V.~Srinivasan, G.~Gao, B.~Holben, J.~Stankovic, E.~Field,
  and K.~Whitehouse.
\newblock The smart thermostat: using occupancy sensors to save energy in
  homes.
\newblock In {\em Proceedings of the 8th ACM Conference on Embedded Networked
  Sensor Systems}, pages 211--224. ACM, 2010.

\bibitem{parson2015dataport}
O.~Parson, G.~Fisher, A.~Hersey, N.~Batra, J.~Kelly, A.~Singh, W.~Knottenbelt,
  and A.~Rogers.
\newblock Dataport and nilmtk: a building data set designed for non-intrusive
  load monitoring.
\newblock 2015.

\bibitem{parson_2012}
O.~Parson, S.~Ghosh, M.~Weal, and A.~Rogers.
\newblock {Non-intrusive load monitoring using prior models of general
  appliance types}.
\newblock In {\em Proceedings of the 26th AAAI Conference on Artificial
  Intelligence}, pages 356--362, Toronto, ON, Canada, 2012.

\bibitem{parson_2014}
O.~Parson, S.~Ghosh, M.~Weal, and A.~Rogers.
\newblock An unsupervised training method for non-intrusive appliance load
  monitoring.
\newblock {\em Artificial Intelligence}, 217:1--19, 2014.

\bibitem{pathak2015acoustic}
N.~Pathak, M.~A. A.~H. Khan, and N.~Roy.
\newblock Acoustic based appliance state identifications for fine-grained
  energy analytics.
\newblock In {\em Pervasive Computing and Communications (PerCom), 2015 IEEE
  International Conference on}, pages 63--70. IEEE, 2015.

\bibitem{pisharoty05thermocoach}
D.~Pisharoty, R.~Yang, M.~M. Newman, and K.~Whitehouse.
\newblock Thermocoach: Reducing home energy consumption with personalized
  thermostat recommendations.
\newblock In {\em Proceedings of the 2nd ACM International Conference on
  Embedded Systems For Energy-Efficient Built Environments}, 2015.

\bibitem{saha2014energylens}
M.~Saha, S.~Thakur, A.~Singh, and Y.~Agarwal.
\newblock Energylens: combining smartphones with electricity meter for accurate
  activity detection and user annotation.
\newblock In {\em Proceedings of the 5th international conference on Future
  energy systems}, pages 289--300. ACM, 2014.

\bibitem{sklearn-feature}
Scikit-learn.
\newblock Feature selection, 2015.

\bibitem{scott2011preheat}
J.~Scott, A.~Bernheim~Brush, J.~Krumm, B.~Meyers, M.~Hazas, S.~Hodges, and
  N.~Villar.
\newblock Preheat: controlling home heating using occupancy prediction.
\newblock In {\em Proceedings of the 13th international conference on
  Ubiquitous computing}, pages 281--290. ACM, 2011.

\end{thebibliography}

\ifCLASSOPTIONcaptionsoff
  \newpage
\fi



%

\end{document}